\documentclass[11pt]{article}

\usepackage[preprint]{acl}

\usepackage{times}
\usepackage{latexsym}

\usepackage[T1]{fontenc}
\usepackage[utf8]{inputenc}
\usepackage{microtype}

\usepackage{graphicx}
\usepackage{amsmath, amsthm, amssymb, amsfonts}
\usepackage{booktabs}
\usepackage{multirow}
\usepackage{enumitem}
\usepackage{nicefrac}
\usepackage{float}
\usepackage{tikz}
\usepackage{pgfplots}
\usepackage{pgf-pie}
\usepackage{tcolorbox}
\tcbuselibrary{listings,skins,breakable}
\pgfplotsset{compat=1.18}

\makeatletter
\renewcommand{\@makefntext}[1]{\noindent\@makefnmark\ #1}
\renewenvironment{abstract}%
  {\begin{center}\large\textbf{\abstractname}\end{center}%
    \begin{list}{}%
      {\setlength{\rightmargin}{0cm}%
        \setlength{\leftmargin}{0cm}}%
      \item[]\ignorespaces%
      \@setsize\normalsize{12pt}\xpt\@xpt
  }%
  {\unskip\end{list}}
\makeatother

\newcommand{\keywords}[1]{}

\newtcolorbox{promptbox}[1][]{
  breakable,
  colback=gray!5,
  colframe=gray!50,
  fonttitle=\bfseries,
  title=#1,
  boxrule=0.5pt,
  arc=2pt,
  left=6pt,
  right=6pt,
  top=4pt,
  bottom=4pt,
}

\title{FinResearchBench II: A Deep Research Benchmark with Consensus-Derived Gold Rubrics for Distinguishing Financial Report Quality}

\author{
  \textbf{Beidi Luan\textsuperscript{1}},
  \textbf{Rui Sun\textsuperscript{1}},
  \textbf{Sinuo Wang\textsuperscript{1,3}\thanks{Work done during internship at Stepfun.}},
  \textbf{Yan Gu\textsuperscript{2}}, \\
  \textbf{Chao Li\textsuperscript{2}},
  \textbf{Zhenliang Xiong\textsuperscript{1,4}}\footnotemark[1],
  \textbf{Jing Li\textsuperscript{1}},
  \textbf{Zuo Bai\textsuperscript{1, 2}}\footnotemark[2]
\\
\\
  \textsuperscript{1}StepFun,
  \textsuperscript{2}FinStep,
  \textsuperscript{3}University of Adelaide,
  \textsuperscript{4}Shanghai Jiao Tong University,
}

\begin{document}
\maketitle
\footnotetext[2]{Corresponding author: \texttt{baizuo@stepfun.com}.}

\begin{abstract}
Deep research agents are increasingly used to produce long-form financial reports, yet large-scale evaluation remains bottlenecked by the need for human experts to define and execute high-quality rubrics. We address this problem by proposing a scalable pipeline for generating high-quality rubrics without human experts in the final loop. We build a financial deep research benchmark from 104 real-world user queries and automatically synthesize 14,450 query-specific candidate rubrics from model-generated reports. To justify removing human experts from rubric execution, we compare rubric judgments from three human experts with those from a three-LLM judge panel on a sampled subset, and show that LLM-based evaluation is sufficiently consistent with human evaluation to replace it for large-scale rubric screening, including 98.67\% label-level agreement on jointly unanimous items. We then derive consensus-derived gold rubrics through two filters: a strict consistency filter, which keeps a rubric only if the three LLM judges unanimously agree on every report under the same query, and a distinguishability filter, which keeps a rubric only if it assigns at least one majority-yes and at least one majority-no label across the evaluated systems. This process retains 3,687 consistency-passed rubrics, of which 2,600 remain distinguishable and form the final set of consensus-derived gold rubrics. Using this final rubric set, we obtain clearly differentiated rankings across 10 deep research systems, with item-level pass rates ranging from 58.58\% to 22.23\%. More broadly, because the pipeline removes human-expert execution from rubric generation and evaluation, it is naturally scalable for benchmark evaluation, automatic system comparison, and future studies of evaluation-driven system improvement.

\end{abstract}

\section{Introduction}

Deep research agents are rapidly becoming practical tools for complex financial analysis. Modern systems increasingly combine retrieval augmentation, tool use, iterative planning, and long-context synthesis to gather evidence and compose report-style outputs \cite{lewis2020retrieval,yao2023react,schmidgall2025agent}. Recent surveys and benchmarks document rapid progress across web research, expert report generation, scientific rediscovery, data-centric analysis, and structured-knowledge reasoning \cite{xu2025deepresearchsurvey,zheng2025deepresearcher,du2025deepresearch,han2025deer,wang2026firebench,liu2026datastorm,liu2026kdrbench}. What now limits progress is not report generation alone, but scalable evaluation that is both discriminative and operationally feasible.

This problem is especially acute in finance, where reports are long, data-dense, and highly heterogeneous across queries. Generic long-form metrics fail to capture planning quality, source-grounded reliability, and analytical coverage \cite{que2024hellobench,wu2025longeval,tuohetiyaer2026deepresearcheval}. Rubric- or criteria-based evaluation is therefore increasingly preferred \cite{zheng2023llmasjudge,zhang2025rubricbench,lv2026queryrubrics,sun2025finresearchbench}, but most high-quality pipelines still rely on human experts for rubric design, preference annotation, guidance, verification, or execution \cite{han2025deer,lv2026queryrubrics,sun2025finresearchbench}. That dependence is the core scalability bottleneck.

We argue that \emph{high-quality rubrics need not be produced through a human-expert loop}. Starting from 104 real-world financial queries, we collect 1,040 reports, synthesize 14,450 query-specific candidate rubrics, and first validate a three-LLM judge panel against three human experts on a sampled subset. We then screen the full pool with two filters: a strict consistency filter, which requires unanimous judgments across all 10 reports under a query, and a distinguishability filter, which retains only rubrics assigning both positive and negative majority labels across products. This process yields 3,687 consistency-passed rubrics and 2,600 final \emph{consensus-derived gold rubrics}.

The resulting rubric set supports direct product comparison: pass-rate gaps exceed 36 percentage points across 10 systems. More broadly, removing human experts from the critical execution path makes rubric-based evaluation substantially easier to scale for benchmarking, system comparison, and future evaluation-driven system improvement.

Our main contributions are:
\begin{enumerate}[leftmargin=1.35em, labelsep=0.45em, itemsep=0.25em, topsep=0.35em, parsep=0pt]
    \item \textbf{A deep research benchmark grounded in real financial use cases.} We build a benchmark from 104 real-world financial queries and 1,040 product reports.

    \item \textbf{A scalable, expert-free pipeline for deriving high-quality rubrics.} From 14,450 candidates, we validate a three-LLM judge panel against human experts and then apply consistency and distinguishability filters to derive 2,600 consensus-derived gold rubrics without humans in the final execution loop.

    \item \textbf{A validated and discriminative evaluation signal.} We show that human--LLM consistency is strong enough to support large-scale LLM-based rubric screening, and that the resulting consensus-derived gold rubrics produce stable separation across 10 products.
\end{enumerate}

\begin{figure*}[!t]
    \centering
    \includegraphics[width=0.98\textwidth,trim=10 10 10 10,clip]{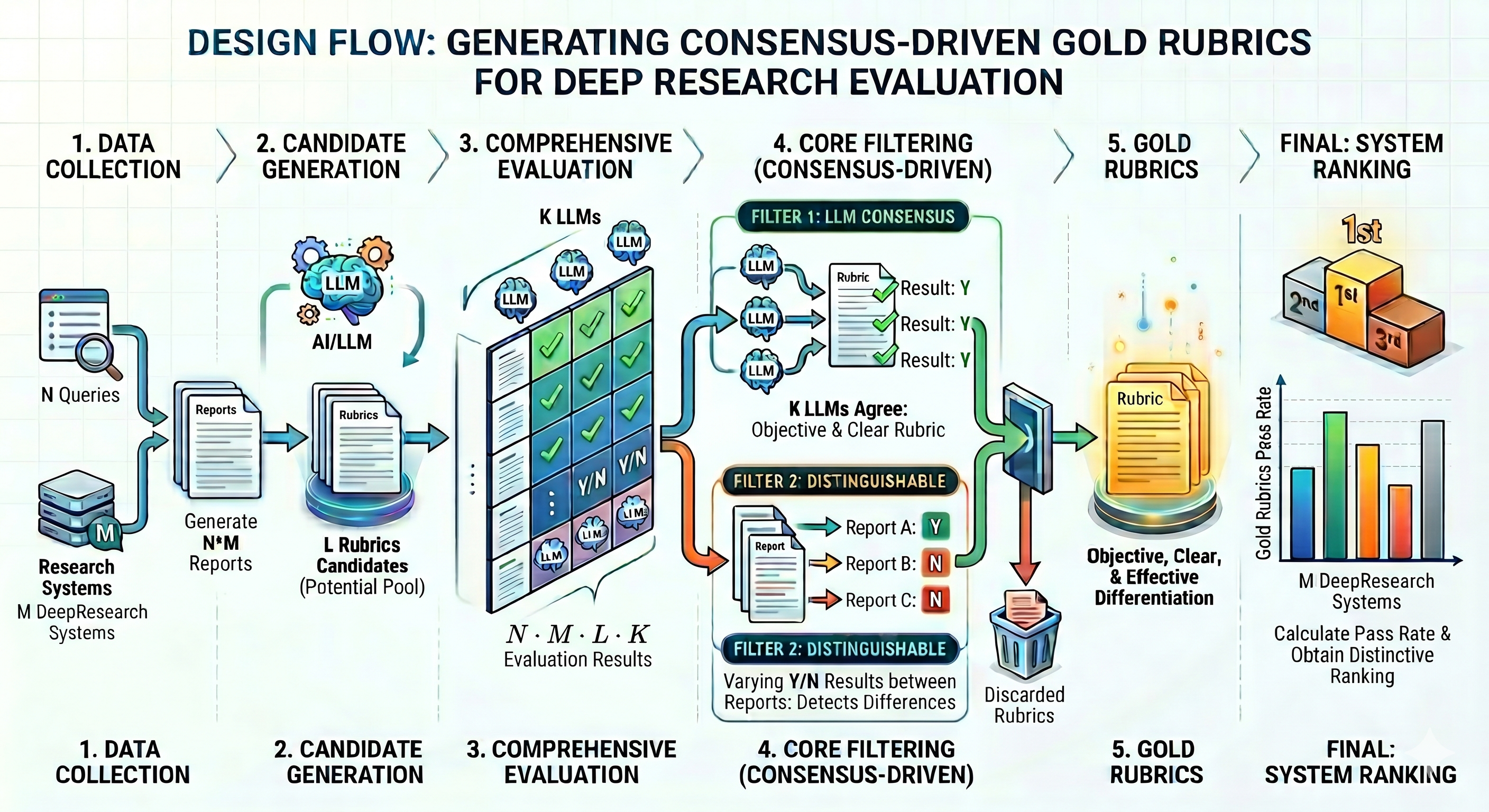}
    \caption{Pipeline for deriving consensus-derived gold rubrics from query-specific candidate rubrics and using the retained rubric set for final system ranking. The figure was generated with Gemini and subsequently reviewed and refined by the authors.}
    \label{fig:design_flow}
\end{figure*}

Figure~\ref{fig:design_flow} summarizes the full pipeline for generating consensus-derived gold rubrics and using them for system ranking.

\section{Related Work}

\subsection{Benchmarks for Deep Research Agents}

Benchmarking has evolved from multi-hop QA \cite{yang2018hotpotqa,trivedi2022musique,ho2020constructing} to action-based agent evaluation \cite{liu2024agentbench,zhuang2023toolqa,deng2023mind2web}, and more recently to deep-research settings involving realistic tool use, report generation, scientific rediscovery, and data-centric analysis \cite{mialon2023gaia,wei2025browsecomp,du2025deepresearch,han2025deer,wang2026firebench,zhong2026draco,liu2026datastorm,liu2026kdrbench}. These works collectively mark a shift toward agentic evaluation of long-horizon research behavior \cite{zhuge2024agent,you2026agentjudge}. Our focus is complementary: not a new task family, but a scalable pipeline for constructing the rubric set used to evaluate report quality.

\subsection{Financial Domain Benchmarks}

Financial evaluation likewise requires domain-specific tasks and criteria. Prior benchmarks cover financial knowledge and reasoning \cite{guo2024fineval,lu2025bizfinbench,lu2025bizfinbenchv2}, agentic retrieval and broader task coverage \cite{wang2025fingaia,hu2025finsearchcomp,xie2024finben}, and real-world scenario evaluation with rubric-based open-ended cases \cite{zhang2026firefin}. Closer to our setting, FinResearchBench \cite{sun2025finresearchbench} introduces a logic-tree-based Agent-as-a-Judge framework tailored to financial research agents, extracting a hierarchical logic tree from each report and applying rule-based metrics over the tree together with LLM-based scoring across 70 questions in 7 task types. Similar specialization appears in legal and medical agents \cite{li2024legalagentbench,tan2026medresearchbench}. These results reinforce that financial report evaluation cannot be reduced to generic open-domain metrics. Our work builds on this direction but shifts the emphasis from a fixed expert-defined evaluation framework to a scalable, expert-free pipeline for deriving query-specific rubrics from the reports themselves.

\subsection{Rubric-Based Evaluation for Long-Form Generation}

Long-form research evaluation increasingly relies on fine-grained criteria rather than BLEU/ROUGE or monolithic LLM judgments. Prior work documents judge biases and the weakness of surface-level proxies \cite{zheng2023llmasjudge,li2024llmasjudge,you2026agentjudge,que2024hellobench,wu2025longeval,bai2024longwriter}. Other work uses rubrics directly \cite{zhang2025rubricbench}, or, in the financial domain, extracts a structural logic tree from each report and derives rule-based metrics on top of an Agent-as-a-Judge system \cite{sun2025finresearchbench}. Deep-Research Eval advocates finer evaluation units and explicit source checks, while query-specific rubric learning uses human preferences to generate task-adaptive supervision \cite{tuohetiyaer2026deepresearcheval,lv2026queryrubrics}. Our work asks a complementary question: how to derive high-quality rubrics at scale without human experts in the final execution loop.

\section{Data Construction}

The construction of a high-quality evaluation benchmark requires data that authentically reflects real-world user needs. Unlike synthetically generated queries or expert-curated hypothetical scenarios, our benchmark is grounded in actual user interactions with a deployed financial deep research service, ensuring ecological validity and practical relevance.

\subsection{Data Collection Methodology}

\begin{figure*}[t]
    \centering
    \begin{tikzpicture}
        \begin{axis}[
            xbar,
            scale only axis,
            width=0.72\textwidth,
            height=8.4cm,
            xmin=0,
            xmax=17.5,
            bar width=4.6pt,
            axis lines*=left,
            tick align=outside,
            symbolic y coords={
                {Individual Stock Identification},
                {Individual Stock Analysis},
                {Stock Screening},
                {Theme/Sector Analysis},
                {Stock Business Query},
                {Stock Trend Prediction},
                {Supply Chain Stock Mining},
                {Fundamental Consultation},
                {Trend \& Trading Advice},
                {Trading Advice},
                {Stock Comparison},
                {Stock Anomaly Analysis},
                {Precious Metals Analysis},
                {Information Query}
            },
            xtick={0,2,4,6,8,10,12,14,16},
            xticklabels={0\%,2\%,4\%,6\%,8\%,10\%,12\%,14\%,16\%},
            xmajorgrids,
            grid style={draw=gray!25},
            xlabel={Share of Queries},
            ytick=data,
            y dir=reverse,
            enlarge y limits=0.03,
            nodes near coords,
            every node near coord/.append style={
                font=\scriptsize,
                anchor=west,
                xshift=2pt,
            },
            point meta=explicit symbolic,
            yticklabel style={
                font=\small,
                text width=3.7cm,
                align=right,
            },
            tick label style={font=\small},
            label style={font=\small},
        ]
        \addplot[
            draw=blue!55!black,
            fill=blue!45,
        ] coordinates {
            (16.0,{Individual Stock Identification}) [16.0\%]
            (14.4,{Individual Stock Analysis}) [14.4\%]
            (12.6,{Stock Screening}) [12.6\%]
            (8.0,{Theme/Sector Analysis}) [8.0\%]
            (7.0,{Stock Business Query}) [7.0\%]
            (7.0,{Stock Trend Prediction}) [7.0\%]
            (6.6,{Supply Chain Stock Mining}) [6.6\%]
            (5.8,{Fundamental Consultation}) [5.8\%]
            (4.4,{Trend \& Trading Advice}) [4.4\%]
            (4.4,{Trading Advice}) [4.4\%]
            (4.0,{Stock Comparison}) [4.0\%]
            (3.6,{Stock Anomaly Analysis}) [3.6\%]
            (3.4,{Precious Metals Analysis}) [3.4\%]
            (2.8,{Information Query}) [2.8\%]
        };
        \end{axis}
    \end{tikzpicture}
    \caption{Distribution of user queries across fine-grained categories.}
    \label{fig:query_distribution}
\end{figure*}
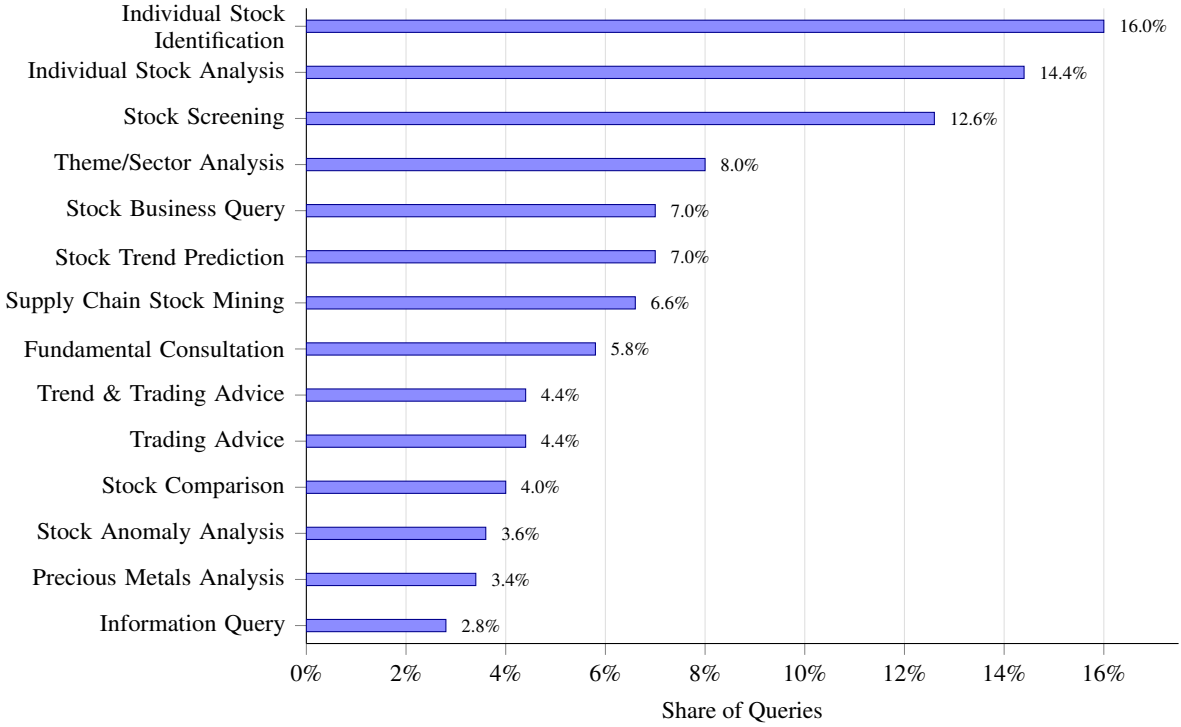

We collected user queries from a deployed financial deep research service during February to March 2026, applying standard privacy protection procedures including anonymization and PII removal. Query analysis revealed seven dominant categories: \textit{Stock Comparison}, \textit{Stock Trend Prediction}, \textit{Fundamental Consultation}, \textit{Stock Trading Advice}, \textit{Theme/Sector Analysis}, \textit{Precious Metals Analysis}, and \textit{Individual Stock Analysis}. Using stratified random sampling across these categories, we constructed a final dataset of 104 queries balancing analytical complexity, temporal diversity, and entity coverage. Figure~\ref{fig:query_distribution} shows the distribution of user queries across fine-grained categories.

\subsection{Rubrics Construction}

The production-grounded nature of our dataset ensures ecological validity, as queries reflect genuine user needs with real-world ambiguity and varying specificity levels. Rather than hand-curating a small fixed rubric inventory, we construct a large query-specific candidate pool directly from model outputs. Concretely, we first collect 10 product reports for each of the 104 queries, yielding 1,040 reports in total. We then prompt LLMs to synthesize binary, query-specific rubric items from these reports, producing 14,450 candidate rubrics overall.

\begin{enumerate}[leftmargin=1.35em, labelsep=0.45em, itemsep=0.25em, topsep=0.35em, parsep=0pt]
    \item \textbf{Candidate Rubric Generation}: For each query, candidate rubrics are generated from the corresponding multi-product report set, with rubric dimensions covering aspects such as comprehensiveness, insight, instruction following, and readability.

    \item \textbf{Human--LLM Validation on a Sampled Subset}: To verify that large-scale rubric screening can be automated, we compare a three-LLM judge panel against three human experts on a sampled subset of rubric items. This step is used as a validation stage rather than as the final large-scale execution mechanism.

    \item \textbf{Consensus-Derived Gold Rubrics}: After validation, the full candidate pool is screened by the three-LLM panel using two filters: a strict consistency filter and a cross-report distinguishability filter. The rubrics that survive both stages form the final set of consensus-derived gold rubrics used for benchmark evaluation.
\end{enumerate}

\section{Evaluation Methodology}
\label{sec:methodology}

\subsection{Expert-Free Query-Specific Rubric Pipeline}

Our methodology is designed around a single objective: deriving high-quality evaluation rubrics without requiring human experts in the final execution loop. Rather than starting from a small, manually curated rubric set, we adopt a query-specific rubric methodology in which evaluation criteria are derived from the reports themselves. For each query, we collect 10 product reports and prompt LLMs to synthesize binary rubric items tailored to that query. Aggregating over all 104 queries produces 14,450 candidate rubrics, which are then screened by an LLM-based pipeline to identify a smaller set of reliable and discriminative rubrics.

\subsubsection{Human-Validated LLM Screening}

Our goal is not to keep human experts in the large-scale execution loop, but to use human judgments once to validate an automated rubric-screening mechanism. We therefore compare three human experts with a three-LLM judge panel on a sampled subset of rubric items. This validation stage supports the operational claim needed by our pipeline: in this benchmark setting, LLM-based rubric evaluation is sufficiently aligned with human evaluation to replace expert execution for large-scale screening.

\subsubsection{Strict Consistency Filter}

After validation, we use the same three-LLM judge panel to evaluate the full candidate pool. For a fixed rubric under a fixed query, each of the 10 product reports receives three binary judgments. A rubric passes the consistency filter only if all three judges agree on every one of the 10 reports. This strict unanimity rule removes rubrics whose execution is not sufficiently objective or reproducible.

\subsubsection{Distinguishability Filter}

Consistency alone is insufficient: a rubric may be perfectly stable while still assigning the same effective judgment to all products. We therefore apply a second filter at the cross-report level. For each consistency-passed rubric, we compute the majority label for each of the 10 product reports. A rubric is considered distinguishable if at least one report receives a majority-yes label and at least one report receives a majority-no label. This removes one-sided rubrics such as always-yes or always-no criteria that are stable but uninformative for product comparison.

\subsubsection{Consensus-Derived Gold Rubrics and Product-Level Pass Rate}

The rubrics that survive both stages are treated as \emph{consensus-derived gold rubrics}. Here, ``gold'' denotes reproducibility and informativeness under the validated consensus pipeline, rather than independent proof of semantic correctness for every item. In our final pipeline, 3,687 rubrics pass the strict consistency filter and 2,600 remain distinguishable. We use this final set as a scalable evaluation signal for comparing products. For a target product $m$, let $\mathcal{G}$ denote the set of consensus-derived gold rubrics and let $y_{m,r} \in \{0,1\}$ indicate whether product $m$ satisfies rubric $r \in \mathcal{G}$. The product-level pass rate is then defined as
\[
\text{PassRate}(m) = \frac{1}{|\mathcal{G}|} \sum_{r \in \mathcal{G}} y_{m,r}.
\]
This metric gives a directly comparable, rubric-grounded measure of report quality across products, and more broadly illustrates how the final set of consensus-derived gold rubrics can serve as an expert-free evaluation signal at scale.

\section{Experiments}

We organize the experiments around four questions. First, is batched rubric evaluation stable within a single judge model when one prompt contains the full rubric set for a report? Second, can unanimity from a three-LLM judge panel serve as a reliable screening signal for rubric executability when compared with human experts? Third, how selective is the proposed two-stage filtering pipeline when applied to the full candidate pool? Fourth, once the final set of consensus-derived gold rubrics is derived, does it separate product quality in a stable and meaningful way?

\subsection{Within-Model Stability of Batched Rubric Evaluation}
\label{sec:exp_within_model}

Our rubric executor does not score one rubric at a time. Instead, for a fixed query--report pair, it places the entire query-specific rubric set into a single prompt and asks the judge model to return a binary decision for every rubric in one shot. This batched design is substantially more efficient, but it raises a natural concern: when a prompt contains around one hundred rubric items, are the resulting labels stable across repeated runs of the same model?

To answer this question, we run a sampled repeated-evaluation diagnostic. For a subset of query--report instances, we execute the same batched prompt three times and compare the resulting rubric labels across runs. For each rubric, we define \emph{within-model consistency} as the majority share among valid yes/no labels across the three rollouts, and we define \emph{flip rate} as the fraction of rubric evaluations in which the repeated runs contain at least one yes/no disagreement.

\begin{table*}[t]
\caption{Within-model stability of batched rubric evaluation on a sampled subset. Full consistency denotes the fraction of rubric evaluations with identical labels in all three runs.}
\label{tab:within_model_stability}
\centering
\begin{tabular}{lcccc}
\toprule
\textbf{Scope} & \textbf{Rubrics} & \textbf{Mean Cons.} & \textbf{Full Cons.} & \textbf{Flip Rate} \\
\midrule
Overall & 6,625 & 0.970 & 90.99\% & 9.01\% \\
Comprehensiveness & 1,671 & 0.971 & 91.38\% & 8.62\% \\
Insight & 2,571 & 0.967 & 89.96\% & 10.04\% \\
Instruction Following & 1,124 & 0.977 & 93.15\% & 6.85\% \\
Readability & 1,259 & 0.969 & 90.63\% & 9.37\% \\
\bottomrule
\end{tabular}
\end{table*}

Table~\ref{tab:within_model_stability} shows that the batched evaluation setup is reasonably stable despite the long rubric lists. The mean within-model consistency is 0.970, the median is 1.0, and 90.99\% of rubric evaluations are fully consistent across all three runs. The rollout error rate is 0.0\%, indicating that instability comes from label variation rather than response-format failures. Because three rollouts are used, every disagreement takes the form of a simple 2--1 split, so the minimum observed consistency is 0.667 rather than complete collapse.

The dimension breakdown is also informative. \textit{Instruction Following} is the most stable category, with a 93.15\% full-consistency rate and a 6.85\% flip rate, while \textit{Insight} is the least stable, with a 89.96\% full-consistency rate and a 10.04\% flip rate. This pattern is intuitively plausible: interpretive rubrics that require deeper synthesis are more variance-sensitive than direct instruction-following checks. Overall, the sampled diagnostic suggests that batching all rubric candidates for a report into one prompt is practically viable.

\subsection{Human--LLM Consistency Validation}
\label{sec:exp_validation}

Before scaling the rubric pipeline to the full benchmark, we validate whether the three-LLM judge panel can support large-scale rubric screening. The validation set compares three human experts with the three-model panel on 4,052 sampled rubric--report items covering all seven query categories and all four rubric dimensions. The human experts are financial analysts with securities-research backgrounds and approximately 10 years of experience each. The LLM panel consists of \texttt{gpt-5.4}, \texttt{claude-opus-4-7}, and \texttt{gemini-3-pro-preview}. Our goal is not to argue that LLM judges universally replace experts, but to test whether LLM unanimity is a dependable criterion for retaining rubrics that are likely to be executed consistently.

\begin{figure}[t]
\centering
\includegraphics[width=0.92\linewidth]{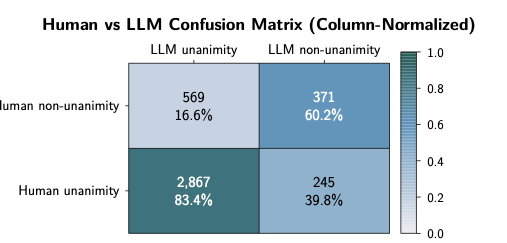}
\caption{Confusion matrix between human unanimity and LLM unanimity on the sampled validation subset. Each cell reports the raw count together with the within-column proportion.}
\label{fig:human_llm_confusion}
\end{figure}

Figure~\ref{fig:human_llm_confusion} shows that the three-LLM panel provides a useful precision-oriented screening signal. Among the 3,436 items for which the LLM panel is unanimous, humans are also unanimous on 2,867 items (83.4\%). These are exactly the kinds of items that our pipeline intends to retain. Conversely, among the 616 items for which the LLM panel is not unanimous, human judges are also non-unanimous on 371 items (60.2\%), indicating that LLM disagreement is informative as a warning signal as well.

We also check whether the two panels assign the same label when both are unanimous. Among the 2,867 jointly unanimous items, label-level agreement is 98.67\% (2,829/2,867; Cohen's $\kappa=0.9733$), with only 38 opposite-label cases. Across all 4,052 validation items, majority-label agreement is 90.42\% ($\kappa=0.8076$). Pairwise agreement further supports the same conclusion: LLM--human agreement (mean agreement 0.848, mean $\kappa=0.694$) exceeds human--human agreement (mean agreement 0.811, $\kappa=0.569$--0.678, mean $\kappa=0.622$). Thus, LLM unanimity is not merely a reproducibility indicator; in this validation subset, it is also a strong proxy for label-level agreement with expert judgments.

The key pattern is that the panel behaves conservatively rather than permissively. It does discard some items that humans judge unanimously, but this is acceptable for our use case because screening is designed to favor precision over recall. In other words, we prefer to keep a smaller set of rubrics whose execution appears objectively reproducible, rather than retain a larger but noisier pool. The confusion matrix therefore supports the specific operational claim needed for this paper: LLM unanimity is a credible basis for conservative rubric retention at benchmark scale.

\subsection{Filtering Statistics}
\label{sec:exp_filtering}

We next apply the rubric pipeline to the full candidate pool of 14,450 query-specific rubrics. The two filters encode complementary requirements for a benchmark rubric: it should be \emph{stable} under judge execution and \emph{informative} for product comparison.

\begin{table*}[t]
\caption{Rubric filtering statistics under the three-LLM pipeline.}
\label{tab:filtering_stats}
\centering
\begin{tabular}{lccc}
\toprule
\textbf{Stage} & \textbf{Count} & \textbf{Stage Rate} & \textbf{Overall Rate} \\
\midrule
Candidate rubrics & 14,450 & --- & 100.00 \\
Pass consistency filter & 3,687 & 25.52 & 25.52 \\
Pass distinguishability filter & 2,600 & 70.52 & 17.99 \\
\bottomrule
\end{tabular}
\end{table*}

Table~\ref{tab:filtering_stats} shows that the strict consistency filter is highly selective: only 3,687 rubrics (25.52\%) survive the requirement that \texttt{gpt-5.4}, \texttt{claude-opus-4-7}, and \texttt{gemini-3-pro-preview} agree on all 10 reports under the same query. This large reduction confirms that raw rubric generation produces many items that are plausible on the surface but not sufficiently reproducible in execution. The second-stage distinguishability filter retains 2,600 of these 3,687 rubrics (70.52\%), yielding a final overall retention rate of 17.99\% from the original candidate pool.

The retained-rubric statistics further clarify the role of each stage. At the query level, the average rubric count drops from 138.94 in the candidate pool to 35.45 after consistency filtering, and then to 25.0 after both filters. The final retained count ranges from 6 to 47 rubrics per query, indicating that even aggressive filtering still leaves a non-trivial evaluation set for every query. Among the 1,087 consistency-passed rubrics that fail distinguishability, 1,056 are always-yes and only 31 are always-no. This asymmetry suggests that unconstrained rubric generation tends to overproduce broadly affirmative checks that many reports satisfy. As a result, distinguishability is not a minor cleanup step; it is essential for turning an executable rubric pool into a benchmark that can actually separate products.

This comparison also functions as a filter-stage ablation. If we remove the distinguishability filter and use all 3,687 consistency-passed rubrics, the product ranking is preserved relative to the final consensus-derived set (Spearman $\rho=1.0$), but the item-level top-to-bottom pass-rate spread shrinks from 36.35 percentage points to 25.63 percentage points. In other words, consistency alone yields stable but flatter scores, whereas distinguishability removes mostly always-yes criteria and increases the discriminative range of the benchmark.

\subsection{Consensus-Derived Gold Rubrics Evaluation}
\label{sec:exp_rubrics}

We evaluate 10 commercial deep research products in total using both the 3,687 consistency-passed rubrics and the final set of 2,600 consensus-derived gold rubrics obtained after distinguishability filtering. Table~\ref{tab:rubrics_results} reports the nine published products, while one held-out Internal System is described separately in the text. For each rubric set, we report two complementary metrics: Item-level Pass Rate, defined as the proportion of rubrics on which the product receives a positive label; and Query-level Macro Pass Rate, defined by first computing the pass rate within each query and then averaging across the 104 queries with equal weight. The former reflects aggregate performance over the retained rubric pool, while the latter controls for query-level imbalance in rubric counts.

\begin{table*}[t]
\caption{Product-level pass rates on the nine published deep research products under the consistency-passed rubric set and the final consensus-derived rubric set. Query-level Macro Pass Rate first computes a pass rate within each query and then averages over the 104 queries.}
\label{tab:rubrics_results}
\centering
\footnotesize
\setlength{\tabcolsep}{4pt}
\begin{tabular*}{\linewidth}{@{\extracolsep{\fill}}lcccc@{}}
\toprule
\textbf{Product} & \multicolumn{2}{c}{\textbf{Consistency-Passed}} & \multicolumn{2}{c}{\textbf{Consensus Gold}} \\
\cmidrule(lr){2-3}\cmidrule(lr){4-5}
 & \textbf{Item Pass} & \textbf{Query Macro} & \textbf{Item Pass} & \textbf{Query Macro} \\
\midrule
Kimi & 69.95 & 71.08 & 58.58 & 59.48 \\
Doubao & 66.37 & 67.21 & 53.50 & 53.38 \\
Gemini & 66.23 & 67.08 & 53.31 & 52.48 \\
ChatGPT & 61.57 & 62.46 & 46.69 & 46.82 \\
Qwen & 61.54 & 62.42 & 46.65 & 46.72 \\
Metasota & 57.26 & 58.31 & 40.58 & 40.09 \\
AutoGLM & 52.48 & 53.18 & 33.81 & 32.87 \\
Minimax & 47.30 & 48.08 & 26.46 & 25.72 \\
Perplexity & 44.32 & 45.06 & 22.23 & 23.11 \\
\bottomrule
\end{tabular*}
\end{table*}

Table~\ref{tab:rubrics_results} reveals two complementary patterns. First, pass rates on the 3,687 consistency-passed rubrics are uniformly higher than those on the final 2,600 consensus-derived gold rubrics. This is expected: distinguishability filtering removes many one-sided rubrics, especially always-yes items, thereby making the retained set more demanding. For reference, the held-out Internal System obtains 64.36/65.36 on the consistency-passed rubric set and 50.65/50.52 on the consensus-derived rubric set (item-level/query-level macro), placing it behind the near-tied Doubao--Gemini tier and ahead of ChatGPT and Qwen. Second, despite this stricter selection, the product ordering among the nine published products remains highly stable. Kimi ranks first under both rubric sets, followed by a near-tied Doubao--Gemini second tier. ChatGPT and Qwen form a middle tier, while Metasota, AutoGLM, Minimax, and Perplexity remain clearly below.

Three conclusions follow from these results. First, the final set of consensus-derived gold rubrics is genuinely more discriminative than the broader consistency-passed set: Kimi's item-level pass rate drops from 69.95\% to 58.58\%, and Perplexity's drops from 44.32\% to 22.23\%, showing that the second filter removes many easy positives and sharpens product separation. Second, the benchmark remains challenging even for the best product. A 58.58\% item-level pass rate means that the leading system still fails more than two-fifths of the final set of consensus-derived gold rubrics, leaving substantial room for future improvement. Third, the query-level macro rankings are nearly unchanged from the item-level rankings under both rubric sets, which suggests that the observed ordering is not driven by a small number of rubric-dense queries but is broadly stable across the benchmark.

The ranking should be interpreted as tiered rather than as a claim of statistically meaningful separation between every adjacent pair. Inter-tier gaps are large, while some within-tier differences are negligible: Doubao and Gemini differ by only 0.19 percentage points in item-level pass rate, and ChatGPT and Qwen are near-tied. We therefore emphasize robust tier-level separation rather than over-interpreting tiny within-tier rank differences.

\subsection{Sensitivity Analyses}
\label{sec:exp_sensitivity}

We run two additional diagnostics to test whether the ranking is an artifact of the judge panel or the distinguishability filter. First, we conduct a leave-one-judge-out analysis by removing \texttt{gpt-5.4}, \texttt{claude-opus-4-7}, or \texttt{gemini-3-pro-preview} individually and recomputing the retained rubric set and product ranking under two-judge unanimity. In all three configurations, the Spearman rank correlation with the full-panel baseline is $\rho=0.988$. The only positional change is a tie-break-level swap between ChatGPT and Qwen, two systems whose baseline scores are effectively tied. For the self-preference concern specifically, removing the Gemini judge leaves the Gemini product unchanged at rank 3; removing the GPT judge causes ChatGPT to drop by one position only because it loses the near-tie with Qwen. These patterns are not consistent with systematic self-preference inflation.

Second, we test whether the distinguishability filter overfits to the evaluated products. For each of the $\binom{10}{3}=120$ splits, we select rubrics using only seven systems and then apply the selected rubrics to rank all 10 systems, including the three held-out systems. Across the 120 splits, the full-ranking Spearman correlation with the baseline averages 0.978, with a minimum of 0.891 and 93.3\% of splits at or above 0.95. For the held-out triples alone, 112 of 120 splits preserve perfectly consistent relative ordering. This indicates that distinguishability acts primarily as a non-triviality filter rather than as an optimization procedure that overfits to a fixed set of products.

\section{Conclusion}

This paper presents a scalable, expert-free pipeline for generating high-quality rubrics for deep research evaluation. Rather than relying on human experts to define and execute consensus-derived gold rubrics at benchmark scale, we begin with 104 real-world financial queries, construct a large pool of query-specific rubric candidates, and use LLM-based evaluation plus consensus-driven filtering to derive a final set of consensus-derived gold rubrics. A key step in this design is the validation of LLM evaluation against human evaluation on a sampled subset, which shows that LLM judges are sufficiently consistent with human experts to replace them for large-scale rubric screening in our setting.

Using the retained rubrics, we obtain clearly differentiated rankings across 10 commercial deep research systems, with substantial separation in pass rates between the strongest and weakest products. This shows that the resulting rubric set is not only executable, but also genuinely discriminative for report-quality comparison.

More broadly, our results suggest that rubric-based evaluation for deep research can move beyond small, manually curated expert pipelines toward a scalable alternative: generate a large candidate pool, validate LLM evaluation against human judgments, and derive high-quality rubrics through consistency and distinguishability filtering. Because this framework removes human-expert execution from the critical path, it is naturally suited to benchmark evaluation, automatic system comparison, and future work on evaluation-driven improvement of deep research systems. Important limitations remain, including temporal sensitivity of financial data, domain specificity, and product-version drift. Future work includes extending the same rubric-generation pipeline to other professional domains and studying how consensus-derived gold rubrics transfer across time and evaluation settings.

\section*{Acknowledgments}
This work utilized generative AI tools for drafting visualizations. Specifically, Fig.~\ref{fig:design_flow} was generated with generative AI assistance and subsequently reviewed and refined by the authors.

This paper was supported by ``2025 Special Program for Supporting Innovative Development in Leading Industries (AI Track) under the High-Quality Industrial Development Initiative'' (Project Name: Finstep Finsmart Intelligent Service Platform; Project ID: 2025-GZL-RGZN-01024) from Shanghai Municipal Commission of Economy and Informatization, Shanghai, China.

\section*{Disclosure of Interests}
The authors have no competing interests to declare that are relevant to the content of this article.

\bibliography{references}

@article{bai2024longwriter,
  author  = {Bai, Y. and Zhang, J. and Lv, X. and others},
  title   = {{Longwriter}: Unleashing 10,000+ word generation from long context {LLMs}},
  journal = {arXiv preprint arXiv:2408.07055},
  year    = {2024}
}

@article{deng2023mind2web,
  author  = {Deng, X. and Gu, Y. and Zheng, B. and others},
  title   = {{Mind2web}: Towards a generalist agent for the web},
  journal = {Advances in Neural Information Processing Systems},
  volume  = {36},
  pages   = {28091--28114},
  year    = {2023}
}

@article{du2025deepresearch,
  author  = {Du, M. and Xu, B. and Zhu, C. and others},
  title   = {{Deepresearch} bench: A comprehensive benchmark for deep research agents},
  journal = {arXiv preprint arXiv:2506.11763},
  year    = {2025}
}

@inproceedings{guo2024fineval,
  author    = {Guo, X. and Xia, H. and Liu, Z. and others},
  title     = {{Fineval}: A chinese financial domain knowledge evaluation benchmark for large language models},
  booktitle = {Proceedings of the 2025 Conference of the Nations of the Americas Chapter of the Association for Computational Linguistics: Human Language Technologies (Volume 1: Long Papers)},
  pages     = {6258--6292},
  year      = {2025}
}

@article{han2025deer,
  author  = {Han, J. and Kim, H. and Lee, C. and others},
  title   = {{Deer}: A comprehensive and reliable benchmark for deep-research expert reports},
  journal = {arXiv e-prints},
  pages   = {arXiv:2512.17776},
  year    = {2025}
}

@inproceedings{ho2020constructing,
  author    = {Ho, X. and Nguyen, A. K. D. and Sugawara, S. and others},
  title     = {Constructing a multi-hop {QA} dataset for comprehensive evaluation of reasoning steps},
  booktitle = {Proceedings of the 28th International Conference on Computational Linguistics},
  pages     = {6609--6625},
  year      = {2020}
}

@article{hu2025finsearchcomp,
  author  = {Hu, L. and Jiao, J. and Liu, J. and others},
  title   = {{Finsearchcomp}: Towards a realistic, expert-level evaluation of financial search and reasoning},
  journal = {arXiv preprint arXiv:2509.13160},
  year    = {2025}
}

@article{lewis2020retrieval,
  author  = {Lewis, P. and Perez, E. and Piktus, A. and others},
  title   = {Retrieval-augmented generation for knowledge-intensive {NLP} tasks},
  journal = {Advances in Neural Information Processing Systems},
  volume  = {33},
  pages   = {9459--9474},
  year    = {2020}
}

@inproceedings{li2024llmasjudge,
  author    = {Li, D. and Jiang, B. and Huang, L. and others},
  title     = {From generation to judgment: Opportunities and challenges of {LLM}-as-a-judge},
  booktitle = {Proceedings of the 2025 Conference on Empirical Methods in Natural Language Processing},
  pages     = {2757--2791},
  year      = {2025}
}

@inproceedings{li2024legalagentbench,
  author    = {Li, H. and Chen, J. and Yang, J. and others},
  title     = {{Legalagentbench}: Evaluating {LLM} agents in legal domain},
  booktitle = {Proceedings of the 63rd Annual Meeting of the Association for Computational Linguistics (Volume 1: Long Papers)},
  pages     = {2322--2344},
  year      = {2025}
}

@article{liu2026datastorm,
  author  = {Liu, S. and Jiang, Y. and Farook, S. and others},
  title   = {{DataSTORM}: Deep Research on Large-Scale Databases using Exploratory Data Analysis and Data Storytelling},
  journal = {arXiv preprint arXiv:2604.06474},
  year    = {2026}
}

@article{liu2026kdrbench,
  author  = {Liu, W. and Li, Z. and Long, B. and others},
  title   = {Towards Knowledgeable Deep Research: Framework and Benchmark},
  journal = {arXiv preprint arXiv:2604.07720},
  year    = {2026}
}

@article{liu2024agentbench,
  author  = {Liu, X. and Yu, H. and Zhang, H. and others},
  title   = {{Agentbench}: Evaluating {LLMs} as agents},
  journal = {arXiv preprint arXiv:2308.03688},
  year    = {2023}
}

@article{lu2025bizfinbench,
  author  = {Lu, G. and Guo, X. and Zhang, R. and others},
  title   = {{Bizfinbench}: A business-driven real-world financial benchmark for evaluating {LLMs}},
  journal = {arXiv preprint arXiv:2505.19457},
  year    = {2025}
}

@article{lu2025bizfinbenchv2,
  author  = {Guo, X. and Zhang, R. and Lu, G. and others},
  title   = {{{BizFinBench} v2}: A Unified Dual-Mode Bilingual Benchmark for Expert-Level Financial Capability Alignment},
  journal = {arXiv preprint arXiv:2601.06401},
  year    = {2026}
}

@article{lv2026queryrubrics,
  author  = {Lv, C. and Zhou, J. and Zhao, W. and others},
  title   = {Learning Query-Specific Rubrics from Human Preferences for {DeepResearch} Report Generation},
  journal = {arXiv preprint arXiv:2602.03619},
  year    = {2026}
}

@inproceedings{mialon2023gaia,
  author    = {Mialon, G. and Fourrier, C. and Wolf, T. and others},
  title     = {{Gaia}: a benchmark for general {AI} assistants},
  booktitle = {The Twelfth International Conference on Learning Representations},
  year      = {2023}
}

@article{que2024hellobench,
  author  = {Que, H. and Duan, F. and He, L. and others},
  title   = {{Hellobench}: Evaluating long text generation capabilities of large language models},
  journal = {arXiv preprint arXiv:2409.16191},
  year    = {2024}
}

@inproceedings{sun2025finresearchbench,
  title={Finresearchbench: A logic tree based agent-as-a-judge evaluation framework for financial research agents},
  author={Sun, Rui and Bai, Zuo and Zhang, Wentao and Zhang, Yuxiang and Zhao, Li and Sun, Shan and Qiu, Zhengwen},
  booktitle={Proceedings of the 6th ACM International Conference on AI in Finance},
  pages={656--664},
  year={2025}
}

@inproceedings{schmidgall2025agent,
  author    = {Schmidgall, S. and Su, Y. and Wang, Z. and others},
  title     = {Agent laboratory: Using {LLM} agents as research assistants},
  booktitle = {Findings of the Association for Computational Linguistics: EMNLP 2025},
  pages     = {5977--6043},
  year      = {2025}
}

@article{tan2026medresearchbench,
  author  = {Tan, S. and Tian, Z.},
  title   = {{MedResearchBench}: A Multi-Domain Benchmark for Evaluating {AI} Research Agents on Clinical Medical Research},
  journal = {medRxiv},
  pages   = {2026.03.30.26349749},
  year    = {2026}
}

@article{trivedi2022musique,
  author  = {Trivedi, H. and Balasubramanian, N. and Khot, T. and others},
  title   = {{{MuSiQue}}: Multihop Questions via Single-hop Question Composition},
  journal = {Transactions of the Association for Computational Linguistics},
  volume  = {10},
  pages   = {539--554},
  year    = {2022}
}

@article{tuohetiyaer2026deepresearcheval,
  author  = {Tuohetiyaer, Y. and Zhu, Y. and Hu, Y. and others},
  title   = {{{Deep-Research Eval}}: An Automated Framework for Assessing Quality and Reliability in Long-Form Reports},
  journal = {Applied Sciences},
  volume  = {16},
  number  = {5},
  pages   = {2546},
  year    = {2026}
}

@article{wang2026firebench,
  author  = {Wang, Z. and Bai, F. and Luo, Z. and others},
  title   = {{{FIRE-Bench}}: Evaluating Agents on the Rediscovery of Scientific Insights},
  journal = {arXiv preprint arXiv:2602.02905},
  year    = {2026}
}

@article{wang2025fingaia,
  author  = {Zeng, L. and Lou, F. and Wang, Z. and others},
  title   = {{{FinGAIA}}: A Chinese Benchmark for {AI} Agents in Real-World Financial Domain},
  journal = {arXiv preprint arXiv:2507.17186},
  year    = {2025}
}

@article{wei2025browsecomp,
  author  = {Wei, J. and Sun, Z. and Papay, S. and others},
  title   = {{Browsecomp}: A simple yet challenging benchmark for browsing agents},
  journal = {arXiv preprint arXiv:2504.12516},
  year    = {2025}
}

@article{wu2025longeval,
  author  = {Wu, S. and Li, Y. and Qu, X. and others},
  title   = {{Longeval}: A comprehensive analysis of long-text generation through a plan-based paradigm},
  journal = {arXiv preprint arXiv:2502.19103},
  year    = {2025}
}

@article{xie2024finben,
  author  = {Xie, Q. and Han, W. and Chen, Z. and others},
  title   = {{Finben}: A holistic financial benchmark for large language models},
  journal = {Advances in Neural Information Processing Systems},
  volume  = {37},
  pages   = {95716--95743},
  year    = {2024}
}

@article{xu2025deepresearchsurvey,
  author  = {Xu, R. and Peng, J.},
  title   = {A comprehensive survey of deep research: Systems, methodologies, and applications},
  journal = {arXiv preprint arXiv:2506.12594},
  year    = {2025}
}

@inproceedings{yang2018hotpotqa,
  author    = {Yang, Z. and Qi, P. and Zhang, S. and others},
  title     = {{{HotpotQA}}: A dataset for diverse, explainable multi-hop question answering},
  booktitle = {Proceedings of the 2018 Conference on Empirical Methods in Natural Language Processing},
  pages     = {2369--2380},
  year      = {2018}
}

@inproceedings{yao2023react,
  author    = {Yao, S. and Zhao, J. and Yu, D. and others},
  title     = {{{ReAct}}: Synergizing Reasoning and Acting in Language Models},
  booktitle = {International Conference on Learning Representations (ICLR)},
  year      = {2023}
}

@article{you2026agentjudge,
  author  = {You, R. and Cai, H. and Zhang, C. and others},
  title   = {{{Agent-as-a-Judge}}},
  journal = {arXiv preprint arXiv:2601.05111},
  year    = {2026}
}

@article{zhang2025rubricbench,
  author  = {Zhang, Q. and Zhou, J. and Wang, Y. and others},
  title   = {{{RubricBench}}: Aligning Model-Generated Rubrics with Human Standards},
  journal = {arXiv preprint arXiv:2603.01562},
  year    = {2026}
}

@article{zhang2026firefin,
  author  = {Zhang, X. and Wu, H. and Guo, J. and others},
  title   = {{{FIRE}}: A Comprehensive Benchmark for Financial Intelligence and Reasoning Evaluation},
  journal = {arXiv preprint arXiv:2602.22273},
  year    = {2026}
}

@article{zheng2023llmasjudge,
  author  = {Zheng, L. and Chiang, W. L. and Sheng, Y. and others},
  title   = {Judging {LLM}-as-a-judge with {MT-Bench} and Chatbot Arena},
  journal = {Advances in Neural Information Processing Systems},
  volume  = {36},
  pages   = {46595--46623},
  year    = {2023}
}

@inproceedings{zheng2025deepresearcher,
  author    = {Zheng, Y. and Fu, D. and Hu, X. and others},
  title     = {{{Deepresearcher}}: Scaling deep research via reinforcement learning in real-world environments},
  booktitle = {Proceedings of the 2025 Conference on Empirical Methods in Natural Language Processing},
  pages     = {414--431},
  year      = {2025}
}

@article{zhong2026draco,
  author  = {Zhong, J. and Zhang, H. and Southern, C. and others},
  title   = {{{DRACO}}: a Cross-Domain Benchmark for Deep Research Accuracy, Completeness, and Objectivity},
  journal = {arXiv preprint arXiv:2602.11685},
  year    = {2026}
}

@article{zhuang2023toolqa,
  author  = {Zhuang, Y. and Yu, Y. and Wang, K. and others},
  title   = {{Toolqa}: A dataset for {LLM} question answering with external tools},
  journal = {Advances in Neural Information Processing Systems},
  volume  = {36},
  pages   = {50117--50143},
  year    = {2023}
}

@article{zhuge2024agent,
  author  = {Zhuge, M. and Zhao, C. and Ashley, D. and others},
  title   = {{{Agent-as-a-judge}}: Evaluate agents with agents},
  journal = {arXiv preprint arXiv:2410.10934},
  year    = {2024}
}

\clearpage
\appendix

\section{Rubrics Evaluation Prompt}
\label{sec:appendix_prompts}

The following prompt template is used for LLM-based report evaluation against binary rubrics:

\begin{promptbox}[Rubrics Evaluation Prompt]
\ttfamily\small
You are a professional financial research report reviewer. Please evaluate whether the given research report satisfies the specified scoring criterion.

\textbf{\#\# User Query}\\
\{query\}

\textbf{\#\# Research Report}\\
\{report\}

\textbf{\#\# Scoring Criterion}\\
\{rubric\}

\textbf{\#\# Scoring Requirements}\\
Please carefully read the research report and determine whether it satisfies the above scoring criterion.
\begin{itemize}
    \item If the report satisfies the criterion, answer ``Yes''
    \item If the report does not satisfy the criterion, answer ``No''
\end{itemize}

Please only answer ``Yes'' or ``No'', no explanation needed.

\textbf{\#\# Your Answer:}
\end{promptbox}

\newpage
\section{Query Examples}

Table~\ref{tab:query_examples} lists translated examples from each category; the original user queries in the dataset are in Chinese.

\begin{table}[H]
\caption{Translated Query Examples by Category}\label{tab:query_examples}
\centering
\footnotesize
\setlength{\tabcolsep}{3pt}
\begin{tabular}{p{0.30\linewidth}p{0.58\linewidth}}
\toprule
\textbf{Category} & \textbf{Example Query} \\
\midrule
Stock Trend Prediction & What is the latest analysis on the rise or fall of Haixia Innovation stock tomorrow? \\
\midrule
Fundamental Consultation & What are the reasons for the decline in Longbai Group's net profit? \\
\midrule
Stock Trading Advice & Does Aluminum Corporation of China have long-term investment value? \\
\midrule
Theme/Sector Analysis & A comprehensive review of AI application sub-sectors. \\
\midrule
Stock Comparison & Which is more suitable for investment, Huitai Medical or Xinmai Medical? \\
\midrule
Precious Metals Analysis & When is the right time for gold price correction? \\
\midrule
Individual Stock Analysis & Conduct an in-depth analysis of Luxshare Precision, examining whether its recent investment logic can become a market hotspot, and provide investment recommendations with target price. \\
\bottomrule
\end{tabular}
\end{table}

\end{document}